\documentclass[letterpaper, 10pt, conference]{ieeeconf}  
\IEEEoverridecommandlockouts                              
\usepackage{fourier}
\usepackage{color}
\usepackage{graphicx}
\usepackage{url}
\usepackage{amsmath}
\usepackage{wrapfig}
\usepackage{xspace}

\usepackage{array}
\usepackage{subcaption}

\usepackage{multicol}
\usepackage{svg}
\usepackage{breakcites}

\usepackage[T1]{fontenc}
\usepackage{times}
\makeatletter
\let\NAT@parse\undefined
\makeatother
\usepackage{hyperref}

\hypersetup{
    colorlinks = true,
    citecolor = blue,
}
\usepackage[nameinlink]{cleveref}
\usepackage{orcidlink}

\title{\LARGE \bf
Scalable and Efficient Hierarchical Visual Topological Mapping
}

\author{
    Saravanabalagi Ramachandran$^{1}$ \orcidlink{0000-0001-6543-5345},
    Jonathan Horgan$^{2}$,
    Ganesh Sistu$^{2}$ and
    John McDonald$^{1}$ \orcidlink{0000-0001-9225-673X}
    \thanks{
        $^{1}$Saravanabalagi Ramachandran and John McDonald are with Lero - the Science Foundation Ireland Research Centre and the Department of Computer Science, Maynooth University, Maynooth, Ireland. 
        {\tt\scriptsize \{saravanabalagi.ramachandran, john.mcdonald\}@mu.ie}.
    }
    \thanks{
        $^{2}$Jonathan Horgan and Ganesh Sistu are with Valeo Vision Sytems, Ireland. 
        {\tt\scriptsize \{jonathan.horgan, ganesh.sistu\}@valeo.com}.
    }
}

\date{December 5, 2023}

\begin{document}
\maketitle
\thispagestyle{empty}
\pagestyle{empty}

\begin{abstract}
Hierarchical topological representations can significantly reduce search times within mapping and localization algorithms. Although recent research has shown the potential for such approaches, limited consideration has been given to the suitability and comparative performance of different global feature representations within this context. In this work, we evaluate state-of-the-art hand-crafted and learned global descriptors using a hierarchical topological mapping technique on benchmark datasets and present results of a comprehensive evaluation of the impact of the global descriptor used. Although learned descriptors have been incorporated into place recognition methods to improve retrieval accuracy and enhance overall recall, the problem of scalability and efficiency when applied to longer trajectories has not been adequately addressed in a majority of research studies.
Based on our empirical analysis of multiple runs, we identify that continuity and distinctiveness are crucial characteristics for an optimal global descriptor that enable efficient and scalable hierarchical mapping, and present a methodology for quantifying and contrasting these characteristics across different global descriptors. Our study demonstrates that the use of global descriptors based on an unsupervised learned Variational Autoencoder (VAE) excels in these characteristics and achieves significantly lower runtime. It runs on a consumer grade desktop, up to 2.3x faster than the second best global descriptor, NetVLAD, and up to 9.5x faster than the hand-crafted descriptor, PHOG, on the longest track evaluated (St Lucia, 17.6 km), without sacrificing overall recall performance.
\end{abstract}

\section{Introduction}
\label{Introduction}

Visual loop closure detection is a core component of many vision-based mapping approaches, where the previously visited places are recognized using Content Based Image Retrieval (CBIR) techniques. 
Typically, the image is encoded as a vector of specified length by means of a global descriptor where a similarity metric appropriate to that type of descriptor is utilised for image comparison. Finding the closest match for a query image involves an exhaustive search through all  encoded images using this metric, and consequently, the time taken to search increases linearly as the total number of images in the map increase.



An important approach to reducing this time complexity is to use indexing techniques. In the seminal work of~\cite{sivic2003video}, the authors proposed an approach to search viewpoint invariant region descriptors using inverted file systems and document rankings similar to the ones used in text retrieval systems. 
Following this work,~\cite{nister2006scalable} builds upon popular techniques of indexing descriptors extracted from local regions and use a vocabulary tree trained in an unsupervised fashion that hierarchically quantizes descriptors from image keypoints. 
Although such techniques offer significant advantages over exhaustive searching, they can be infeasible for mapping very large environments.

\begin{figure}
    \vskip 5pt
    \centering
    \includegraphics[width=0.99\columnwidth]{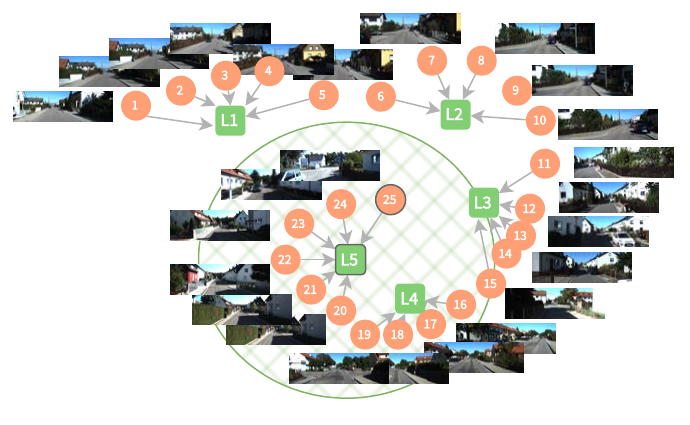}
    \caption{An illustration of Hierarchical Mapping and Localization in the global descriptor embedding space. Locations are shown in green bubbles and their images are shown in orange bubbles connected to them. As $I_{25}$ gets processed, the hatched green circle around $L_5$ shows the reduced search space containing the 3 locations, only within which the loop closing image candidate is searched.}
    \label{fig:hierarchical_mapping}
    \vskip -12pt
\end{figure}

A hierarchical representation of the environment~\cite{topmap_korrapati2014vision, htmap_Garcia-Fidalgo2017}, where images that present a similar appearance are grouped together in nodes, can significantly reduce the search space when finding similar places. As such, the hierarchy helps accelerate the retrieval process by skipping multiple nodes that are not relevant altogether. Although such research have shown the potential for hierarchical matching, limited consideration has been given to the suitability and comparative performance of different feature representations used within these approaches.

We propose to use compact learned global descriptors in hierarchical topological mapping of environments to aggregate sequences of images with similar appearance into location nodes, based on the approach first proposed by~\cite{htmap_Garcia-Fidalgo2017}. 
Many 
learned descriptors with improved retrieval accuracy have been incorporated into place recognition methods to enhance overall recall. In this paper, we focus on addressing the challenges of scalability and efficiency, in particular, when such methods are used on longer trajectories. We show through our evaluation that the use of learned global descriptors is deemed necessary even when hand-crafted global descriptors perform similarly to learned descriptors in terms of recall at $100\%$ precision on benchmark datasets. This is due to learned descriptors' ability to improve efficiency, reduce total runtime, and minimize the total number of relevant locations searched, among other factors.

Our contributions can be summarized as follows:
\begin{itemize}
    \item We extend the Hierarchical Topological Mapping system from~\cite{htmap_Garcia-Fidalgo2017}, perform an in-depth analysis of its components, and make a number of improvements to the underlying implementation, and incorporating learned global descriptors,
    \item We compare hierarchical topological mapping technique with state-of-the-art hand-crafted and learned global descriptors and present results of a comprehensive evaluation of the impact of the global descriptor used,
    \item Through empirical analysis, we identify and define the characteristics of an ideal global descriptor supporting hierarchical matching amenable to scalable and efficient visual localization, and present a methodology for quantifying and contrasting these characteristics, and,
    \item We propose the use of compact learned global descriptors that excel in continuity and distinctiveness characteristics, as an efficient and scalable means for hierarchical topological mapping.
\end{itemize}

\section{Background}
\label{Background}

Hierarchical representations can significantly reduce search times within mapping and localization. Consequently, there have been a number of recent advancements in hierarchical mapping and localization techniques~\cite{hloc_irschara2009structure, hloc_middelberg2014scalable, hloc_sarlin2018leveraging}. A notable state-of-the-art approach is HFNet~\cite{hloc_sarlin_8953492} which follows a hierarchical approach based on a monolithic Convolutional Neural Network (CNN) that simultaneously predicts local features and global descriptors for accurate 6-DoF localization. Although these techniques use hierarchical approaches for more efficient processing, their use of a metric representation makes them intractable to run on longer sequences that are several kilometres long.

There have also been a number of works suitable for very large scale mapping and localization~\cite{fabmap_cummins2008fab_newcollege_citycentre, fabmap2.0_cummins2011appearance, seqslam_milford2012seqslam} without using an explicit metric representation.
More recently,~\cite{htmap_Garcia-Fidalgo2017} proposed an appearance-based approach for topological mapping based on a hierarchical decomposition of the environment where the map aggregates images with similar visual properties together into location nodes, which are represented by means of an average global descriptor and an index of local binary features.

A central decision in the development of each of the above systems is the choice of feature descriptor given its impact on the system's performance. The traditional and hand-crafted global descriptor approaches such as~\cite{sivic2003video, fabmap2.0_cummins2011appearance, galvez2012bags} using Bag of Visual Words (BoW) approach were commonly used in early visual SLAM systems. More recently, the research community has focussed on the use of learned global descriptors given their compelling performance in the field of computer vision in areas such as object recognition, detection, segmentation, and image representation~\cite{resnet_he_2016_7780459, redmon2016yolo, chen2017deeplab, 
prcc_nuimeprn10987}.

In~\cite{arandjelovic2016netvlad} the authors introduced NetVLAD, a generalized differentiable VLAD layer in a CNN trained end-to-end using weak supervision to learn a distance metric based on the triplet loss. Various extensions to NetVLAD have also been proposed, such as~\cite{yu_spe_netvlad_2020_8700608,hausler_patch_netvlad_2021,cai_patch_netvlad_plus_2022,khaliq_multires_netvlad_2022}, to produce patch-level features and/or to capture multi-scale features.

There have also been many attempts to improve performance of image retrieval systems using semantic information.~\cite{smc_toft2018semantic} present a method for scoring the individual correspondences by exploiting semantic information about the query image and the scene to perform a semantic consistency check useful for outlier rejection.~\cite{garg2018lost} proposed the Local Semantic Tensor (LoST) descriptor derived from the convolutional feature maps of a pretrained dense semantic segmentation network. 

More recently,~\cite{sceneCategorization_imvip} use a generalized global descriptor that captures coarse features from the image using an unsupervised convolutional 
Disentangled Inferred Prior Variational Autoencoder (DIPVAE)~\cite{dipvae_kumar2017variational},
to map images to a multi-dimensional latent space. 

As such, several learned image descriptors have emerged in the research field, with a primary focus on capturing finer image features and enhancing retrieval accuracy. However, their ability to scale to visual place recognition over longer trajectories and, in particular, their efficiency in a hierarchical setup has not been adequately evaluated. In this paper we address this issue, proposing the adoption of a global image descriptor that enables image aggregation into locations and minimizes the number of searches of prior locations for achieving scalable and efficient place recognition. Our approach is to extend the hierarchical topological mapping algorithm proposed by~\cite{htmap_Garcia-Fidalgo2017} incorporating the use of learned global descriptors for representing locations. We perform extensive analysis on the impact of the global descriptor on the formation of locations, their discriminability, and the coherence of images within locations, which in turn affects the overall recall and runtime of the algorithm. Through our analysis we identify a set of  required characteristics for feature representation to support scalability and efficiency in hierarchical mapping.

\section{Methodology}
\label{Methodology}

We extend the Hierarchical Topological Mapping (HTMap) algorithm proposed by~\cite{htmap_Garcia-Fidalgo2017} to allow us to evaluate the performance of a variety of learned feature representations. For completeness, we first provide a summary of the relevant elements of the HTMap algorithm here. 


The HTMap algorithm creates a topological map of locations, where each location $L$ is an aggregation of similar images. More formally, $L = \{ I \mid \forall J \in L, \, d(I_\text{gd}, J_\text{gd}) < t_\text{nn} \}$, where the global image descriptor $I_\text{gd}$ encodes a holistic representation of the image, $I$ and $J$ are images, $d$ is the distance function, and $t_\text{nn}$ is the aggregation threshold. Each location also maintains its own location descriptor $L_\text{gd}$ defined as a function of its associated images. This aggregation into a single location descriptor $L_\text{gd}$ creates a two-level hierarchy as shown in \autoref{fig:hierarchical_mapping}, allowing newly processed images to be compared directly with the higher level locations. Loop closing image candidates are then searched only within the set $\{ L \mid 1 - d_n(I_{\text{gd}}, L_\text{gd}) > t_{\text{llc}} \}$, where $t_\text{llc}$ is the location loop closure threshold, and $d_n$ being distance min-max normalized across all locations. In \cite{htmap_Garcia-Fidalgo2017}, Pyramid Histogram of Oriented Gradients (PHOG)~\cite{phog_bosch2007representing} is used as the global descriptor to compute $I_\text{gd}$ with Chi squared distance as the distance function $d$. For a comprehensive explanation of the HTMap algorithm, the reader is advised to refer to~\cite{htmap_Garcia-Fidalgo2017}.

We fork the original HTMap implementation\footnote{available at {\scriptsize\url{https://github.com/emiliofidalgo/htmap}}} and add a number of alterations, enhancements and optimizations to the pipeline for more accurate and efficient localization. 
We fix a number of implementation errors with pre-loading descriptors from disk, $\text{NaN}$ values when calculating distances and normalized similarities, and memory optimization, amongst others. Then, we conduct an in-depth analysis of the performance of individual components constituting the system.

\begin{figure}
    \vskip 5pt
    \centering
    \includegraphics[width=0.99\columnwidth]{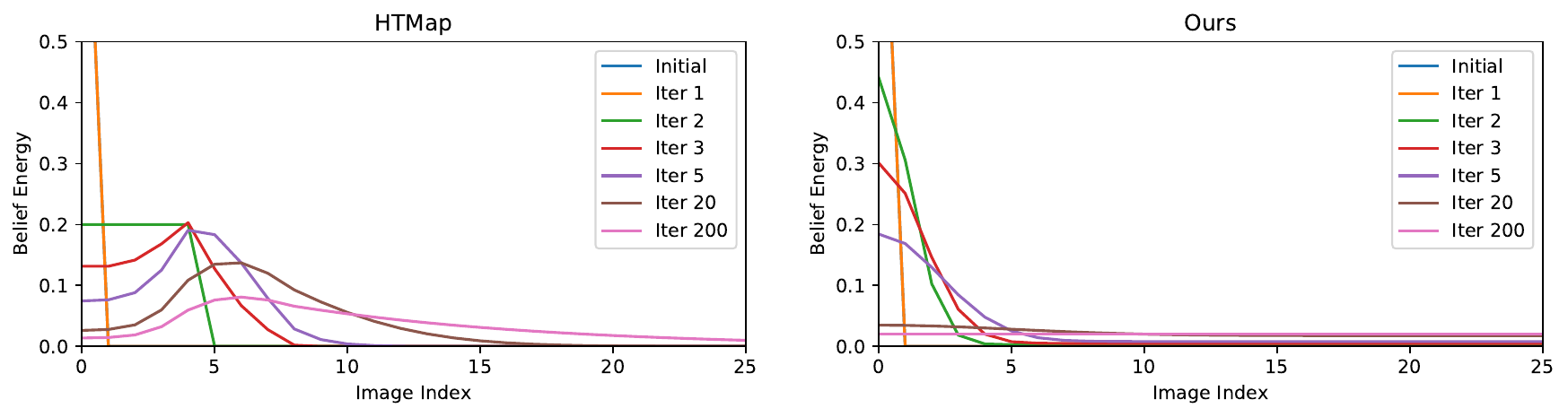}
    \caption{Plots showing beliefs after repeated posterior calculation (involving energy diffusion and normalization)  initialized on a set of beliefs for 100 images. Left: Original HTMap~\cite{htmap_Garcia-Fidalgo2017}: beliefs are not diffused even after 200 iterations, Right: Ours: beliefs are diffused significantly at 20 iterations and completely at 200 iterations.}
    \label{fig:bayes_issue_1}
    \vskip -12pt
\end{figure}

{HTMap} adopts a discrete Bayes filter utilizing an evolution transition model to compute a belief distribution over the set of previously processed images. 
In the original implementation of HTMap, we identified problems in how the energy is accumulated and dissipated over time by the evolution model as shown in \autoref{fig:bayes_issue_1}. In particular, in the absence of measurement updates (i.e. input images), the posterior distribution should eventually distribute its belief uniformly across all poses. Instead, as shown in the figure, a peak is maintained around the early poses of the sequence. We attribute these problems to three main issues: (i) The probability mass adversely accumulates energy around the initial few poses, in addition to gradually shifting away from the first pose. (ii) While calculating the posterior, after dissipating 90\% of the energy to the 8 neighbours, the remaining 10\% of the energy is distributed to all but 8 neighbours, resulting in irregular energy distribution. (iii) As new poses are introduced they are initialized to have zero prior belief, leading to the persistence of accumulated energy around the initial set of poses.

\begin{figure}
    \vskip 5pt
    \centering
    \includegraphics[width=0.99\columnwidth]{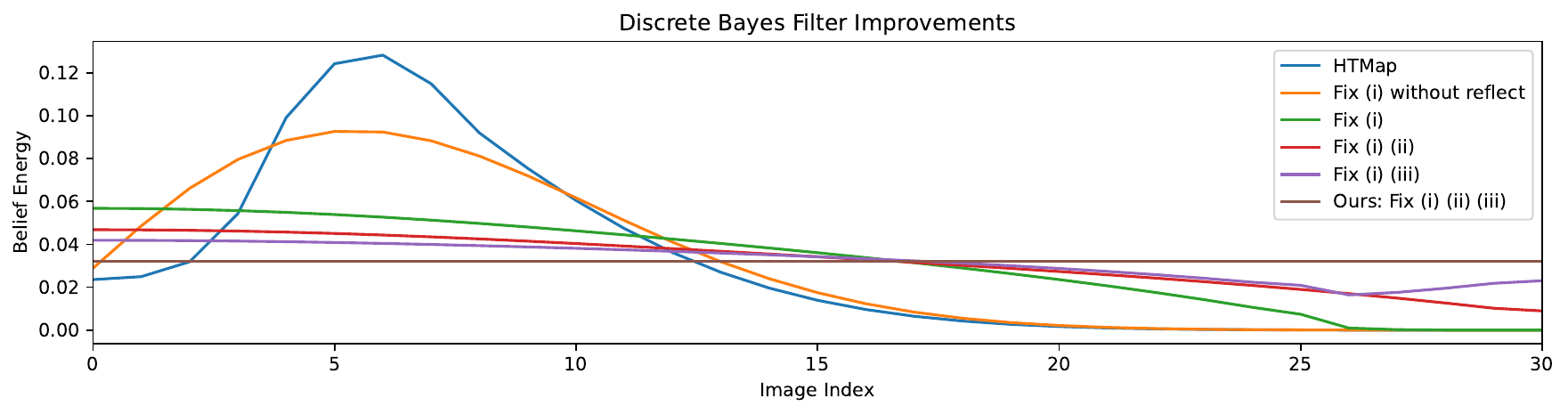}
    \caption{Fixes to discrete Bayes filter ablated: The plot shows the beliefs after processing 30 poses in a  trial started with an initial belief of 1 for the first image and where subsequent poses do not update priors. See \Cref{Methodology} for explanations for fixes (i) to (iii).}
    \label{fig:bayes_improvements}
    \vskip -12pt
\end{figure}

To address these issues, we propose to apply 90\% diffusion using a discrete Gaussian kernel summing to 0.9 with reflect padding to fix (i). To rectify (ii), we distribute remaining 10\% of the energy to all poses. Finally, to fix (iii), we initialise new poses with a value of $1/n$, where $n$ is the number of poses already processed. \autoref{fig:bayes_improvements} ablates the fixes made showing how our model correctly produces a final uniform belief in the absence of measurement updates.

Although the above changes resulted in an improvement in the recall of loop closures in the early stages of mapping session, the overall recall of loop closures exhibited only a marginal increase. On further inspection, we discovered that 
the problems caused by the issues were
masked by the robustness of the local descriptor-based image matching module which returns a significantly higher matching score for true image matches. However, this resilience comes at the expense of a considerable increased runtime, as the module had to process additional locations erroneously updated to have higher beliefs.

More importantly, we adapt the framework to utilize learned global descriptors. For the histogram based global descriptor PHOG, as a new image $I$ is added to the active location, the location descriptor is updated as the mean of active location descriptor $L_\text{gd}$ and the global descriptor of the new image $I_{\text{gd}}$. For learned descriptors, we follow the same approach, however, we 
replace the Chi squared distance with the Euclidean distance. 


For computing the global descriptor $I_\text{gd}$ within {HTMap}, we utilize NetVLAD~\cite{arandjelovic2016netvlad}, LoST~\cite{garg2018lost}, and DIPVAE~\cite{dipvae_kumar2017variational}, which are state-of-the-art in general supervised learned, supervised semantic and unsupervised variational autoencoded latent paradigms, respectively. Several recent variations of NetVLAD, such as Patch-NetVLAD~\cite{hausler_patch_netvlad_2021} and Patch-NetVLAD+~\cite{cai_patch_netvlad_plus_2022}, have been proposed to improve recall performance by providing patch descriptors in addition to the global descriptor. However, our approach only utilizes the global descriptor for image aggregation, and therefore we do not employ these variants. Additionally, while it could be argued that other NetVLAD variants, such as SPE-NetVLAD~\cite{yu_spe_netvlad_2020_8700608} and MultiRes-NetVLAD~\cite{khaliq_multires_netvlad_2022}, could also be used, our primary objective is not to improve recall performance. Instead, our main focus is on assessing the efficacy of a global descriptor in facilitating image aggregation and minimizing location searches for place recognition. As a result, the principal focus of our work is to utilize global descriptors that capture features using different learning approaches.

The integration of such neural networks into the framework presents a set of challenges stemming from variations in programming languages, GPU library constraints, and the potential for increased codebase complexity and tight coupling. 
ONNX~\cite{onnx_bai2019} offers a standardized approach to integrate machine learning models, however, it can be limiting in cases where the neural networks utilize functions and custom operations not supported by the ONNX library. Hence, we implement a mechanism to obtain global descriptors through two approaches: (i) calling an external function for real-time online purposes, and (ii) loading pre-computed descriptors from disk for offline scenarios when available. This ensures that the framework can effectively accommodate diverse models without being tightly bound to any specific implementation. 

\section{Experiments}
\label{Experiments}

\begin{figure*}
    \vskip 5pt
    \begin{minipage}[t]{0.35\textwidth}
    \centering
    \scriptsize
    \begin{tabular}{lcccr}
    \hline
    \textbf{Dataset} & \textbf{\# Imgs} & \textbf{Resolution} & \textbf{Rate} & \textbf{Dist} \\
    & & (px) & (Hz) & (km) \\
    \hline
    City Centre   & 1237             & 1280 × 480          & 0.5           & 2.0               \\
    KITTI 00      & 4541             & 1241 × 376          & 10            & 3.7               \\
    KITTI 05      & 2761             & 1226 × 370          & 10            & 2.2               \\
    KITTI 06      & 1101             & 1226 × 370          & 10            & 1.2               \\
    St Lucia      & 21815            & 640 × 480           & 15            & 17.6              \\ \hline
    \end{tabular}
    \captionof{table}{Datasets used for evaluation}
    \label{tab:eval_datasets}
    \end{minipage}
    \quad
    \begin{minipage}[t]{0.6\textwidth}
    \centering
    \scriptsize
    \begin{tabular}{lrcrrr}
    \hline
    \textbf{GDescriptor} & \textbf{Length} $\downarrow$ & \textbf{Type} & \textbf{Device} & \textbf{BSize} $\uparrow$ & \textbf{Compute Time} (s) $\downarrow$ \\ \hline
    PHOG                & 1260            & Handcrafted  & CPU & 16   & 0.005455 | 0.007601       \\
    LoST                & 6144            & Supervised   & GPU & 22   & 0.181668 | 0.231041      \\
    NetVLAD             & 4096            & Supervised   & GPU & 22   & 0.027397 | 0.048229     \\
    NetVLAD Cropped     & \textbf{128}    & Supervised   & GPU & 22   & 0.027907 | 0.049699      \\
    DIPVAE R128         & \textbf{128}    & Unsupervised & GPU & 1500 & 0.000008 | 0.000179      \\
    DIPVAE R64          & \textbf{128}    & Unsupervised & GPU & \textbf{6000} & \textbf{0.000002} | \textbf{0.000040}      \\ \hline
    \end{tabular}
    \captionof{table}{Global Descriptors considered for evaluation. 
    Compute times (per image) are reported for the specified max batch size \textit{BSize} and also for a batch size of 1 separated by a vertical bar.}
    \label{tab:eval_candidates}
    \end{minipage}
    \vskip -12pt
\end{figure*}

In our experimental evaluation we employed the following benchmark datasets, similar to those used in~\cite{htmap_Garcia-Fidalgo2017}: City Centre~\cite{fabmap_cummins2008fab_newcollege_citycentre}, KITTI~\cite{kitti_geiger2012we} (Sequences 00, 05 and 06) and St Lucia~\cite{stlucia_glover2010fab} (Sequence 19-08-09 08:45). For City Centre, images from left and right cameras are horizontally concatenated. For KITTI, RGB images from the middle camera (\textit{Cam 2}) are used. More information about the datasets is given in \Cref{tab:eval_datasets}. 

As part of the in-depth analysis, we utilize OdoViz~\cite{odoviz_9564712} to visualize the image-image ground truth loop closures, superimposed on the trajectory pose information for each dataset. Through this process, we have identified and corrected several erroneous entries in the ground truth loop closure (GTLC) matrices as provided by~\cite{gt_arroyo2014fast} and \cite{htmap_Garcia-Fidalgo2017}. Corrected ground truth loops are made publicly available\footnote{available at {\scriptsize\url{https://github.com/saravanabalagi/htmap_gt_loops}}}.

Furthermore, we analysed the margin criterion $m=10$ frames used to determine a predicted loop closure image as a true positive (TP) in the original HTMap algorithm. In particular, we use OdoViz to perform a manual visual analysis of the datasets and identify that $m=10$ is overly conservative for St Lucia~\cite{stlucia_glover2010fab} dataset given its higher frame-rate and the relatively higher speed of the ego-vehicle.
To ensure 100\% precision, $m=50$ for St Lucia and $m=10$ for all other datasets is used.

\Cref{tab:eval_candidates} provides information regarding the global descriptors used. For PHOG, we use the code extracted from the C++ implementation
provided by authors of~\cite{htmap_Garcia-Fidalgo2017}. For NetVLAD, the MATLAB implementation of the off-the-shelf {\textit{VGG16 + NetVLAD + whitening}} model pretrained on Pitts30k~\cite{pitts_torii2013visual} dataset
provided by the authors of~\cite{arandjelovic2016netvlad} is used. We additionally use a NetVLAD 128 variant where we crop and L2 normalize the NetVLAD 4096 dimensional descriptor to 128 dimensions. 
For LoST, we use the MATLAB implementation of RefineNet~\cite{lin2017refinenet} pretrained on CityScapes~\cite{cordts2016cityscapes} and Python implementation of the LoST descriptor from RefineNet embeddings provided by the authors of~\cite{garg2018lost}. 
For DIPVAE, the PyTorch implementation with the architecture proposed by the authors pretrained on Oxford Robotcar dataset used in~\cite{sceneCategorization_imvip} is used. We use R128 and R64 DIPVAE variants, both with an embedding length of 128, pre-trained on 128x128 and 64x64 RGB images respectively.



Experiments were carried out on a PC equipped with an Intel i9-9900K (8 cores @3.60 GHz) with 32 GiB DDR4 RAM and a single Nvidia GeForce RTX 2080 Ti with 11 GiB DDR6 VRAM. The HTMap algorithm is run with a 32 byte LDB local descriptor and $t_\text{inliers}$ set to 75, 80, 80, 125, and 75 inliers for City Centre, KITTI 00, 05, 06 and St Lucia, respectively to obtain 100\% precision i.e. zero false loop closures. Other default settings as suggested in~\cite{htmap_Garcia-Fidalgo2017} are used and HTMap is run multiple times by varying the $t_{nn}$ parameter to obtain a different number of locations on each dataset.

\begin{figure*}
    \centering
    \begin{subfigure}{\linewidth}
        \centering
        \makebox[20pt]{\raisebox{50pt}{\rotatebox[origin=c]{90}{\tiny City Centre}}}%
        \includegraphics[width=0.31\textwidth]{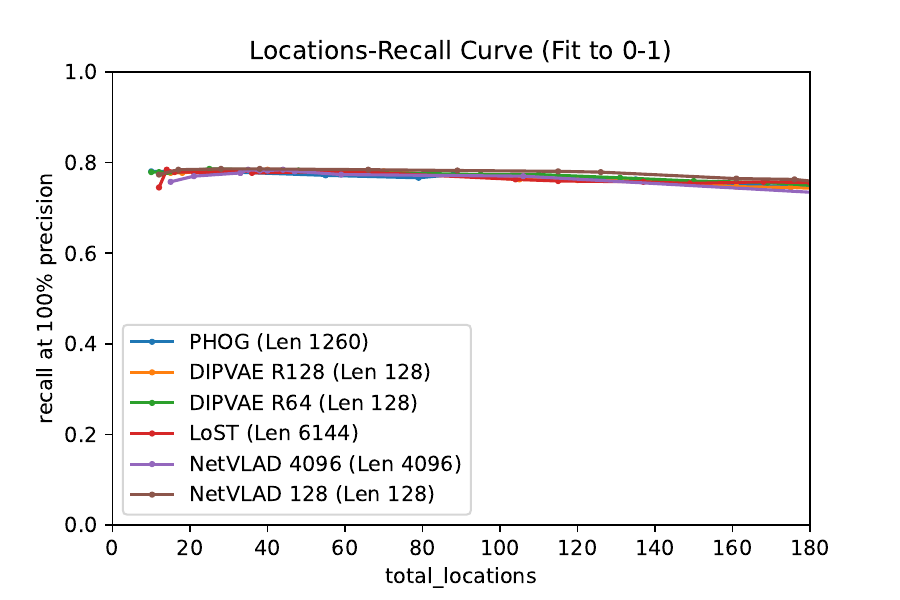}
        \includegraphics[width=0.31\textwidth]{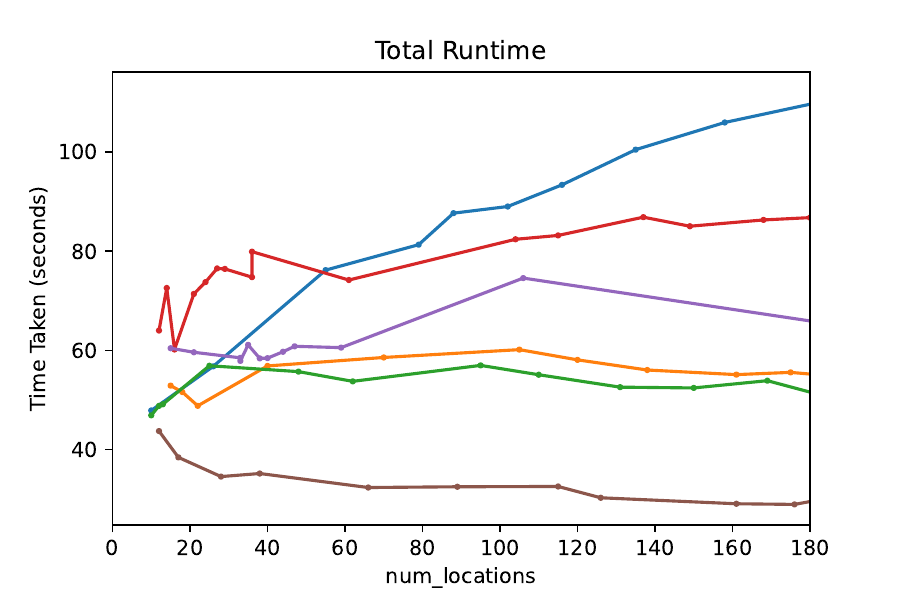}
        \includegraphics[width=0.31\textwidth]{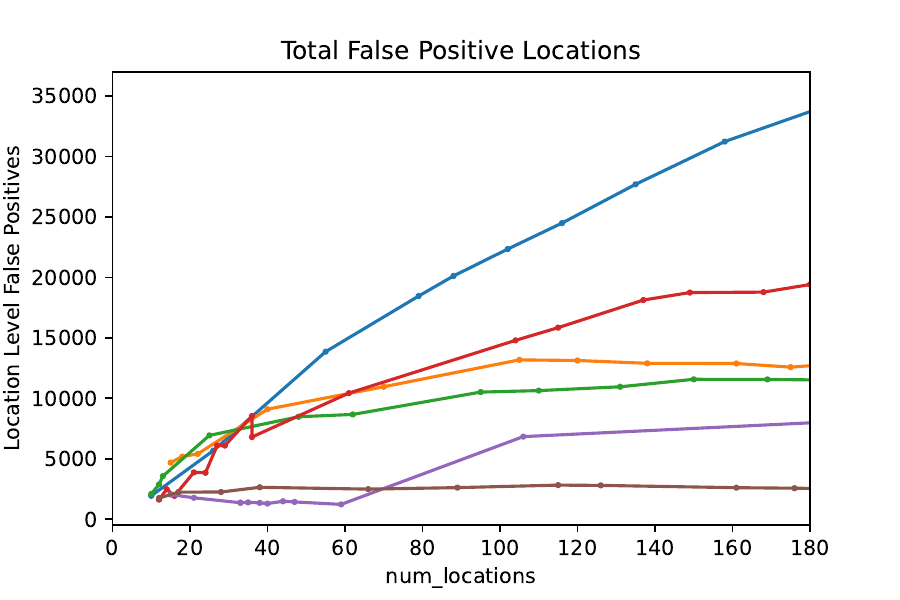}
    \end{subfigure}
    
    \bigskip
    \begin{subfigure}[t]{\linewidth}
        \centering
        \makebox[20pt]{\raisebox{50pt}{\rotatebox[origin=c]{90}{\tiny KITTI 00}}}%
        \includegraphics[width=0.31\textwidth]{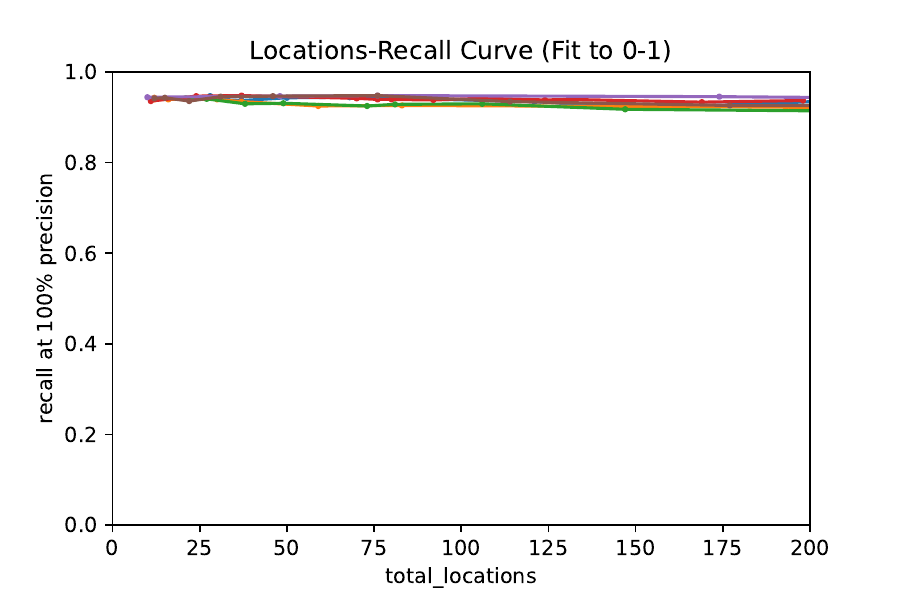}
        \includegraphics[width=0.31\textwidth]{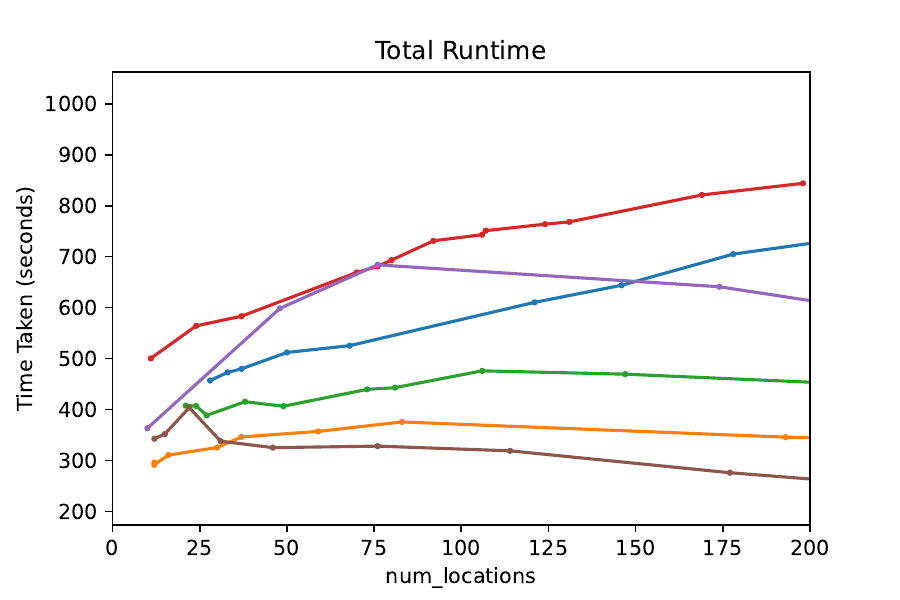}
        \includegraphics[width=0.31\textwidth]{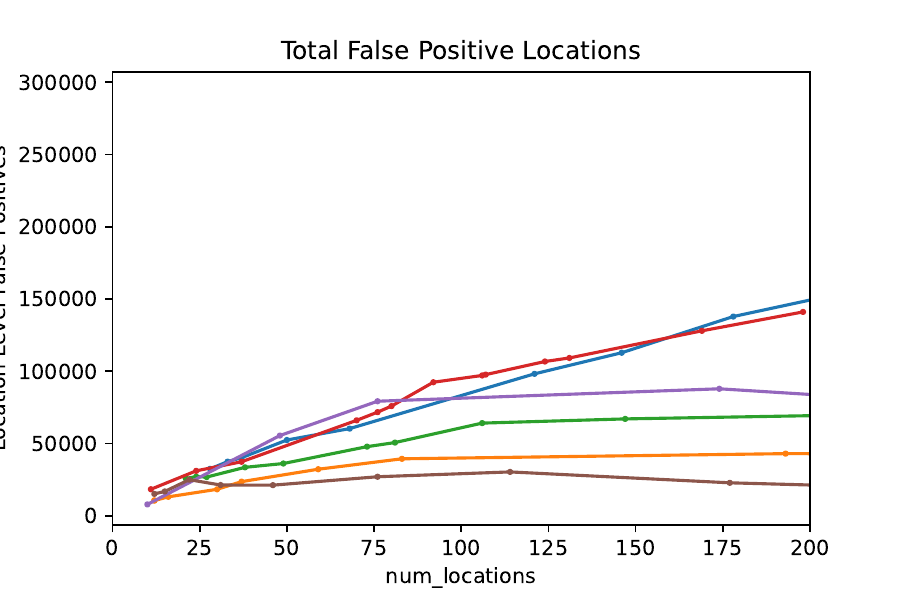}
    \end{subfigure}
    
    \bigskip
    \begin{subfigure}[t]{\linewidth}
        \centering
        \makebox[20pt]{\raisebox{50pt}{\rotatebox[origin=c]{90}{\tiny KITTI 05}}}%
        \includegraphics[width=0.31\textwidth]{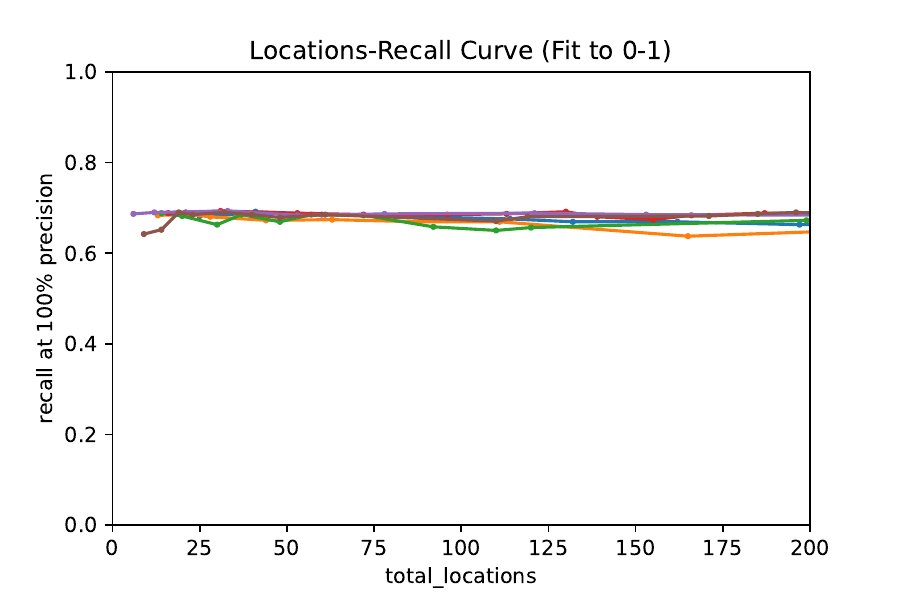}
        \includegraphics[width=0.31\textwidth]{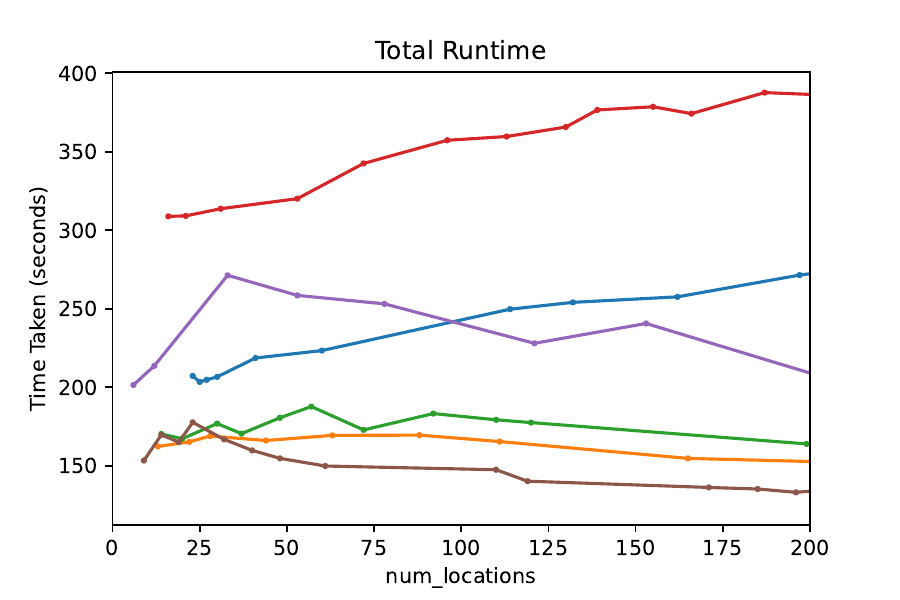}
        \includegraphics[width=0.31\textwidth]{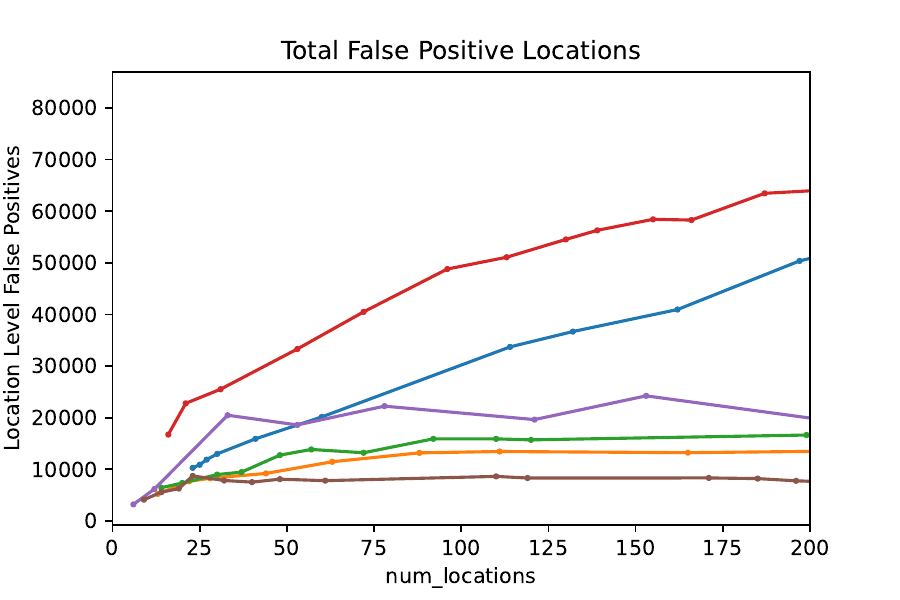}
    \end{subfigure}
    
    \bigskip
    \begin{subfigure}{\linewidth}
        \centering
        \makebox[20pt]{\raisebox{50pt}{\rotatebox[origin=c]{90}{\tiny KITTI 06}}}%
        \includegraphics[width=0.31\textwidth]{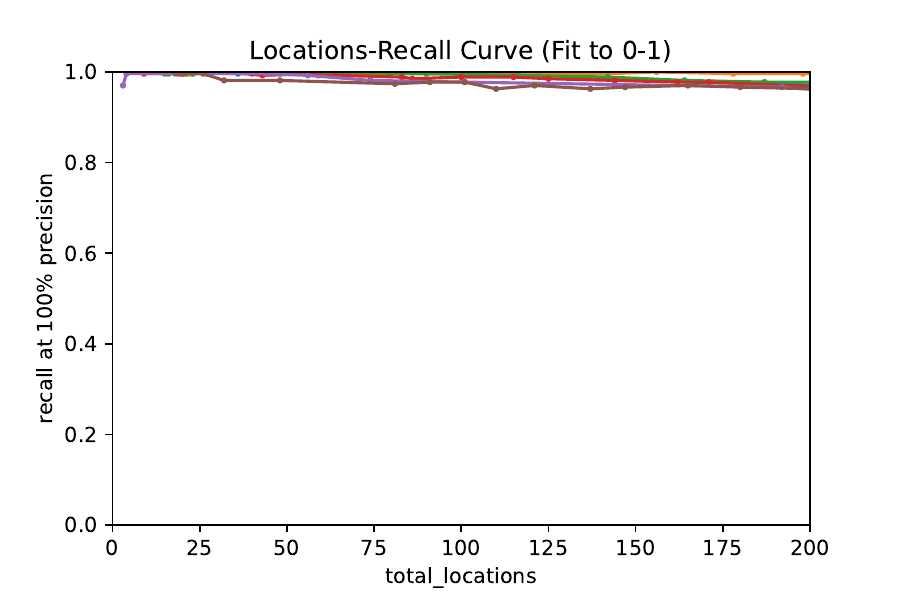}
        \includegraphics[width=0.31\textwidth]{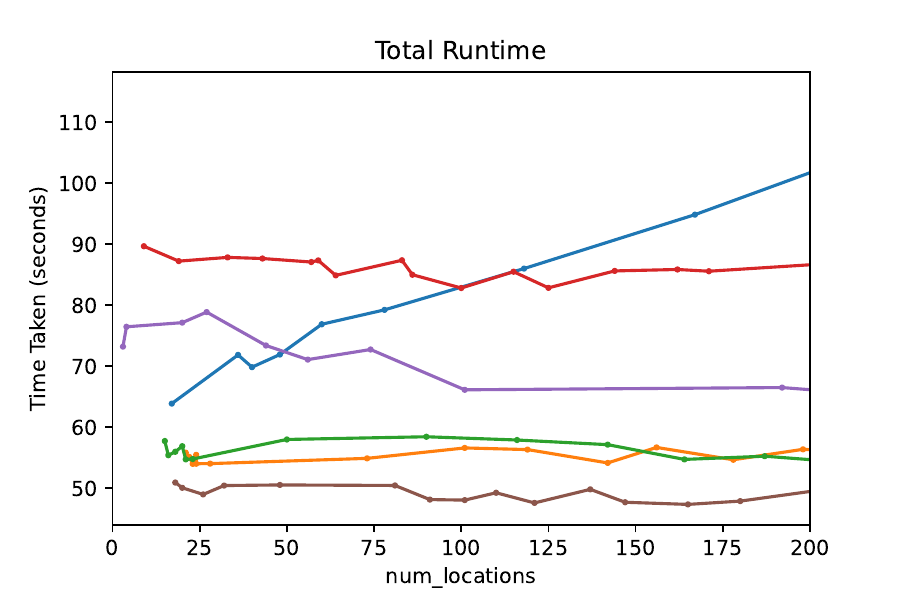}
        \includegraphics[width=0.31\textwidth]{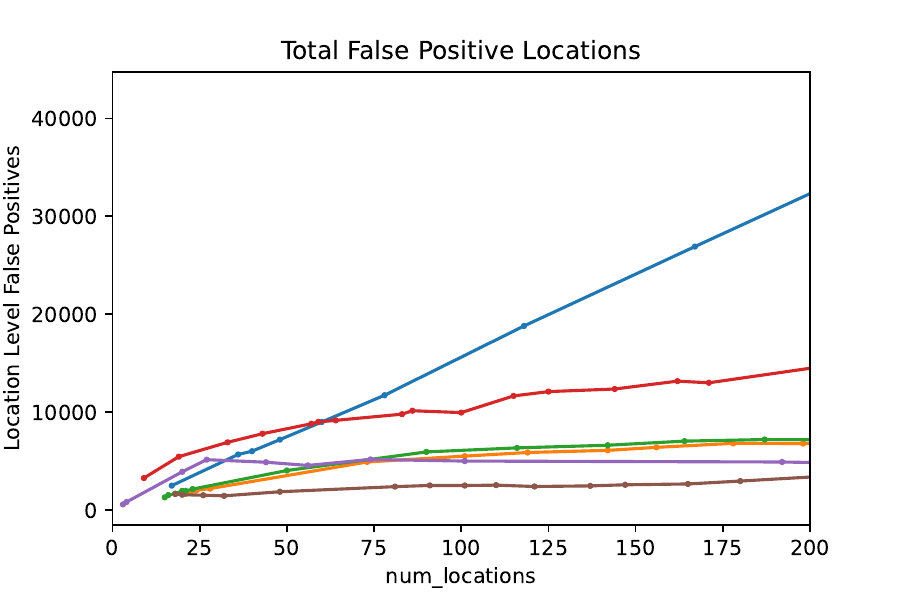}
    \end{subfigure}
    
    \bigskip
    \begin{subfigure}{\linewidth}
        \centering
        \makebox[20pt]{\raisebox{50pt}{\rotatebox[origin=c]{90}{\tiny St Lucia}}}%
        \includegraphics[width=0.31\textwidth]{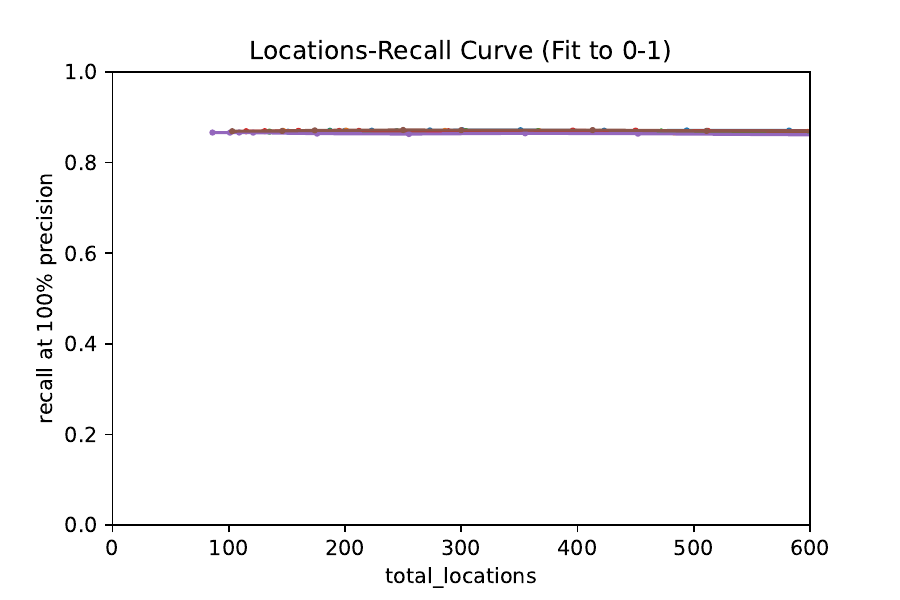}
        \includegraphics[width=0.31\textwidth]{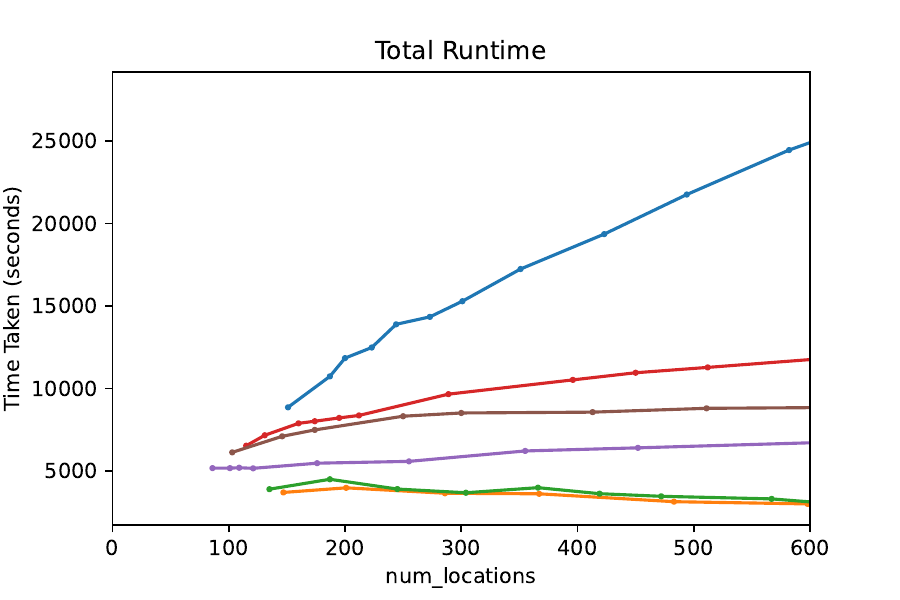}
        \includegraphics[width=0.31\textwidth]{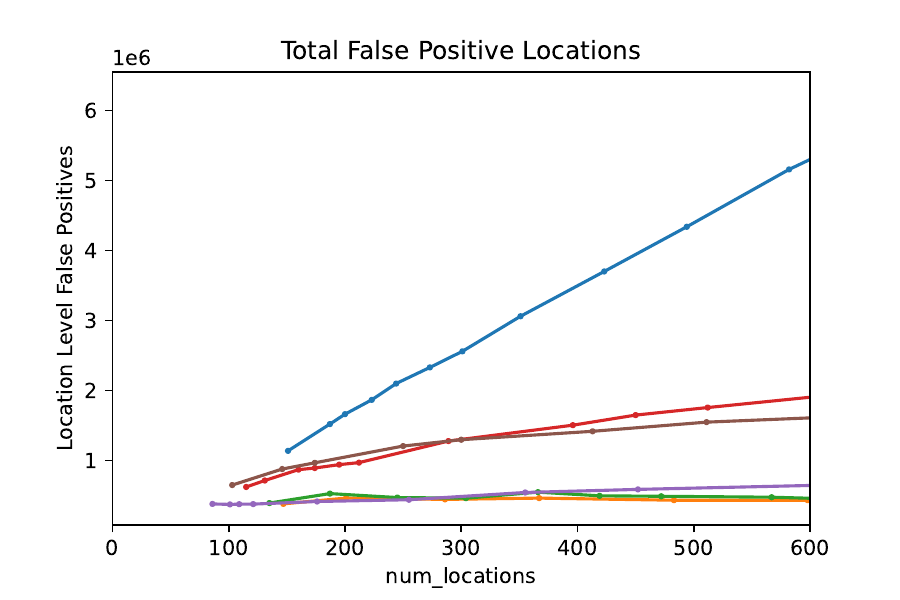}
    \end{subfigure}
    \caption{Results of evaluation of 6 global descriptors across 5 benchmark datasets. Each dot represents one run with the line showing a series of runs of the corresponding global descriptor (legend is given only in the first graph to avoid clutter) on the respective dataset (shown on the left margin). Column 1 highlights that DIPVAE (both R64 and R128) variants maintain the same performance, achieving recall values similar to that of other descriptors, whilst being significantly more compact and faster to compute. Vector plots presented here are best viewed zoomed on a high resolution screen.}
    \label{fig:eval_results}
\end{figure*}

The min and max $t_\text{nn}$ are chosen such that they yield an upper bound of 250 locations (700 for St Lucia) and a lower bound of 10 locations (100 for St Lucia) respectively. 
The bounds and the scale of $t_\text{nn}$ varies significantly for different global descriptors. Hence, for each global descriptor, we plot recall at 100\% precision and the total time taken against the number of locations obtained in the map, see \autoref{fig:eval_results}. Each curve in the graph corresponds to a global descriptor and each data-point corresponds to one run of the algorithm with a specific $t_\text{nn}$. We also plot the number of False Positive Location Candidates (FPLC) proposed, which is an inverse measure of the number of relevant locations searched. As such, more FPLC proposals require searching within indices of multiple locations and would significantly contribute to increased runtimes.

Additionally, to measure the loss of recall due to the use of the hierarchical representation, we run the same set of experiments with ground truth loop closure location proposals. 
This results in zero FPLC and hence runs with the least total runtime. Usually a drop in recall is expected as a result of not proposing correct loop closing ground truth locations, however, in some cases, we also noticed a slight increase in recall.
We find that certain loop closure misses cause the formation of new locations amenable to further loop closures (similar to fragmentation shown in \autoref{fig:odoviz_locations_compared}), resulting in a small increase in recall. Hence, we do not use the loss in recall using ground truth location proposals as a metric to measure performance.

Further, the descriptor compute time for each global descriptor is recorded and compared. To ensure fair comparison, Python implementation\footnote{Implementations available at: \\
PHOG: {\scriptsize\url{https://github.com/saravanabalagi/phog}}, \\
NetVLAD: {\scriptsize\url{https://github.com/Nanne/pytorch-NetVlad}}, \\
LoST: {\scriptsize\url{https://github.com/DrSleep/refinenet-pytorch}}} of the descriptor models were used.
\Cref{tab:eval_candidates} shows batch and individual compute times measured for each global descriptor along with the inference device used. Batch compute times are measured with maximum batch size \textit{BSize} possible (limited by memory capacity and/or CPU Cores) and the mean compute time per image is reported. We also report compute times for the single image inference (i.e., batch size of 1). The first image from the City Centre evaluation dataset was used for inference to measure the compute times of all models.


\section{Discussion and Inference}
\label{Discussion_and_Inference}

\begin{table}
\vskip 5pt
\centering
\begin{tabular}{rrrr}
\hline
\multicolumn{1}{l}{\textbf{$t_\text{nn}$}} & \multicolumn{1}{l}{\textbf{Recall}} & \multicolumn{1}{l}{\textbf{Locations}} & \multicolumn{1}{l}{\textbf{Image Loops}} \\ \hline
1.35  & 0.7772  & 24  & 436 \\
1.40  & 0.7790  & 21  & 437 \\
\textbf{1.43}  & \textbf{0.7843}  & \textbf{14}  & \textbf{440} \\
\textbf{1.50}  & \textbf{0.7790}  & \textbf{16}  & \textbf{437} \\
1.60  & 0.7451  & 12  & 418 \\ \hline
\end{tabular}
\caption{Table showing an example where a higher value of $t_\text{nn}$ 1.50 yields more locations than that of $t_\text{nn}$ 1.43, in rows 3 and 4 (highlighted in bold) respectively for LoST global descriptor on City Centre dataset}
\label{tab:more_locations_anomaly}
\vskip -12pt
\end{table}

The recall at 100\% precision only changes negligibly among the evaluated candidates as we can observe from Column 1 of \autoref{fig:eval_results}, showing that all global descriptors propose sufficient loop closing location candidates without suffering any considerable drop in recall. However, the total time taken varies substantially and can be seen to be correlated in most of the cases with that of the total number of FPLC proposed. From the experimental results produced on different datasets, as the number of locations increases, the runtime, and the number of FPLC proposed thereof, increase linearly for PHOG, almost linearly for LoST although less than that of PHOG, while NetVLAD and DIPVAE descriptors consistently show near-constant runtime with only a very small or negligible proportional increase in FPLC. Also, in both batch and single image descriptor compute times, DIPVAE gives the least compute time amongst all, followed by PHOG. On the other hand, NetVLAD and LoST require comparatively longer compute times making them less suitable for realtime inference on autonomous vehicles.

Upon closer examination, we observed a significantly imbalanced distribution of images across locations in runs with longer runtime. Maps with fewer locations comprising a very large number of images within locations diminish the advantage of the hierarchical approach and impair search efficiency, whereas an increased number of locations, each comprising only a very few images, leads to more location searches. Therefore, achieving optimal performance requires avoiding both overpopulated and underpopulated locations. The max intra-cluster distance $t_\text{nn}$ affects the location cluster cohesion and separation, i.e., determines how many locations are formed and how populated the locations are. A small value for $t_\text{nn}$ usually results in a map with a large number of total locations and larger $t_\text{nn}$ value yields fewer locations.  

\begin{figure}
    \vskip 5pt
    \centering
    \includegraphics[width=\columnwidth]{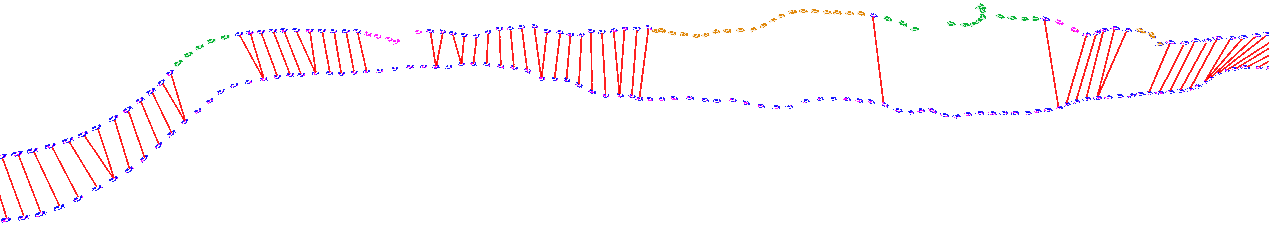} \\
    \vspace{6pt}
    \includegraphics[width=\columnwidth]{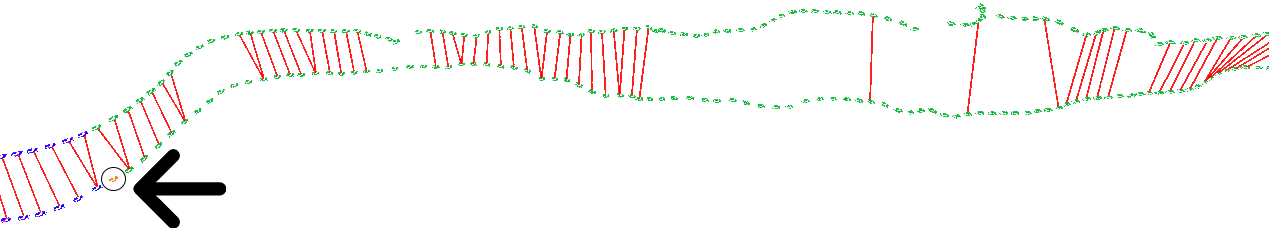}
    \caption{Screenshots of a section of a City Centre trajectory visualized in OdoViz~\cite{odoviz_9564712} showing more fragmentation (different colours along the trajectory) for a higher value of $t_\text{nn}$ 1.50 (top) compared to $t_\text{nn}$ 1.43 (bottom). The car moves from left to right, time is represented in z-axis with newer traversal presented higher up, poses that belong to the same location are in the same colour, and image loop closure proposals are shown in red. The missed loop closure proposal due to the creation of a new location (orange) in the bottom image is circled.}
    \label{fig:odoviz_locations_compared}
    \vskip -12pt
\end{figure}

However, this balance is also influenced by other factors. The location-image aggregation affects the loop closure proposals and the loop closures affects the image aggregation dynamics for subsequent images, which in-turn affect further loop closures. Consequently, it is possible that a slightly higher value of $t_\text{nn}$ yields significantly more locations due to loop closures as shown in \autoref{tab:more_locations_anomaly}. Further, paradoxically, the generation of certain novel locations and subsequent loop closure misses can lead to an increase in overall recall in certain instances, which can be attributed to the dynamic hierarchical structure as shown in \autoref{fig:odoviz_locations_compared}.

\begin{figure}
    \vskip 5pt
    \centering
    \includegraphics[width=0.49\columnwidth]{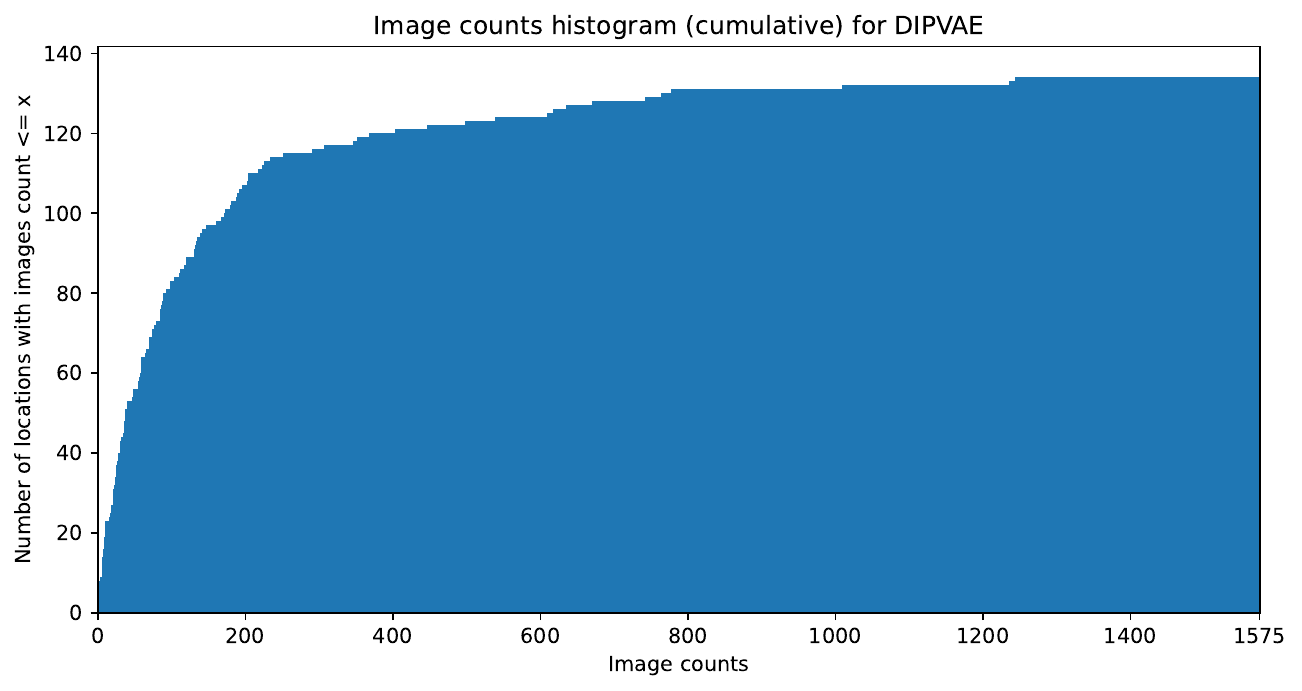}
    \includegraphics[width=0.49\columnwidth]{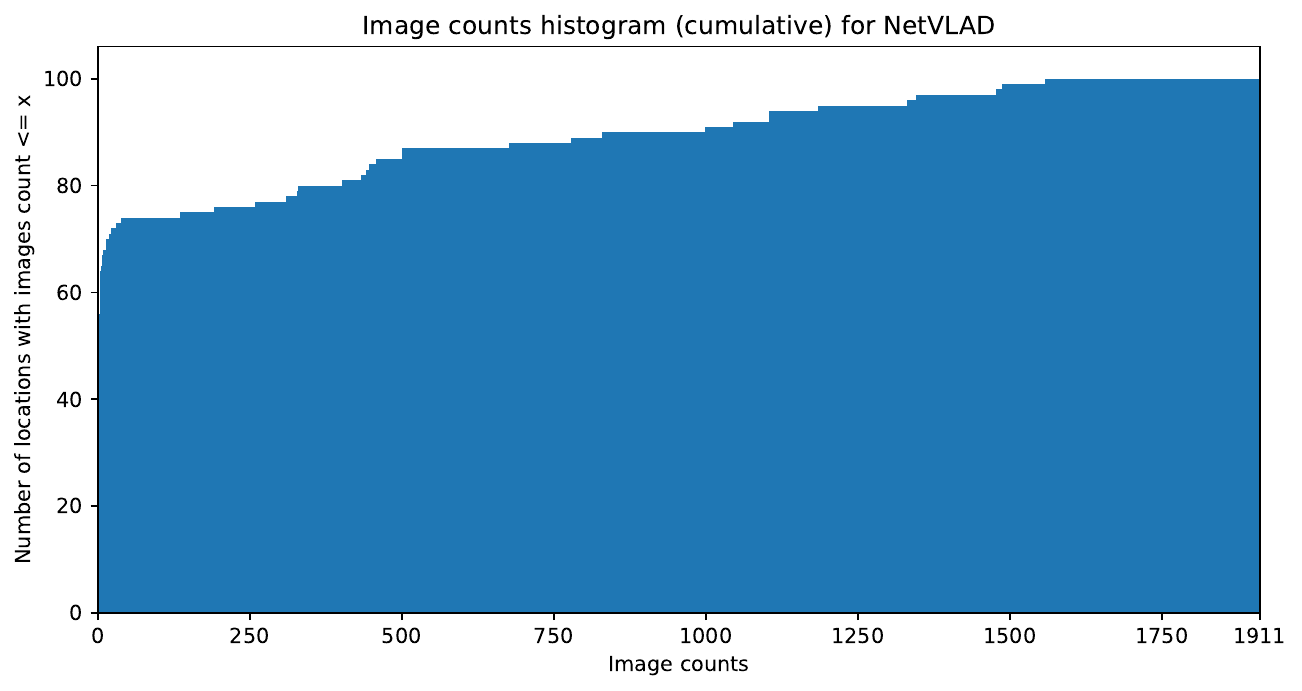}
    \caption{Cumulative histogram of image counts for St Lucia dataset (21,815 images in total) for DIPVAE (left) and NetVLAD (right), illustrating the number of locations that possess an image count less than or equal to the given value of image count. Despite having fewer total number of locations, NetVLAD has 65 out of 101 locations with 5 or fewer images, compared to 11 out of 135 for DIPVAE.}
    \label{fig:img_count_hist}
    \vskip -12pt
\end{figure}

Furthermore, our analysis reveals that the characteristics of the global descriptor has a very significant influence on the association of images with the locations. The generation of highly distinct descriptors for consecutive frames and images of physically proximal regions by a global descriptor results in a large number of locations, each containing only a few images. 
Although this may only lead to a negligible or slight increase in FPLC, it still results in significantly longer runtime (NetVLAD 4096 in Row 5, Column 2 and 3, in \autoref{fig:eval_results}) 
due to the sheer number of location searches triggered. We refer to this characteristic as continuity and posit that it is one of the significant factors in determining runtime. We determine continuity by computing the ratio of the number of locations containing less than $t_\text{ci}$ images to the total number of locations, where $t_\text{ci}$ is directly proportional to the frame rate of the camera and inversely proportional to the average speed of the car. \autoref{fig:img_count_hist} presents a histogram depicting the number of images in all locations in St Lucia. As shown, for NetVLAD, due to its low continuity, the majority of locations are sparsely populated with fewer than 5 images, resulting in excessive fragmentation.
Conversely, a global descriptor that produces similar global descriptors for images that are not physically nearby can lead to the unnecessary search of numerous known locations 
for place recognition, resulting in a large number of FPLC proposals and consequently increasing the runtime. We refer to this characteristic as distinctiveness, which is also crucial in determining runtime. We quantify distinctiveness by computing the inverse of the number of FPLC proposed by the global descriptor. The FPLC proposals for PHOG, as shown in Column 3 of \autoref{fig:eval_results}, increase significantly as the total number of locations increases, indicating its low distinctiveness.

Based on our empirical analysis, we hypothesize that an ideal global descriptor should possess the following characteristics:

\begin{itemize}
    \item \textbf{Continuity}: Descriptor distance should gradually decrease as frames change continuously, resulting in smooth changes in similarity across space
    \item \textbf{Distinctiveness}: Descriptor distance between images from different regions should be significantly larger than the distance to its consecutive frames and images from similar regions.
\end{itemize}

\begin{figure}
    \vskip 5pt
    \centering
    \begin{subfigure}[t]{\columnwidth}
        \centering
        \begin{subfigure}[t]{0.46\columnwidth}
            \includegraphics[width=\columnwidth]{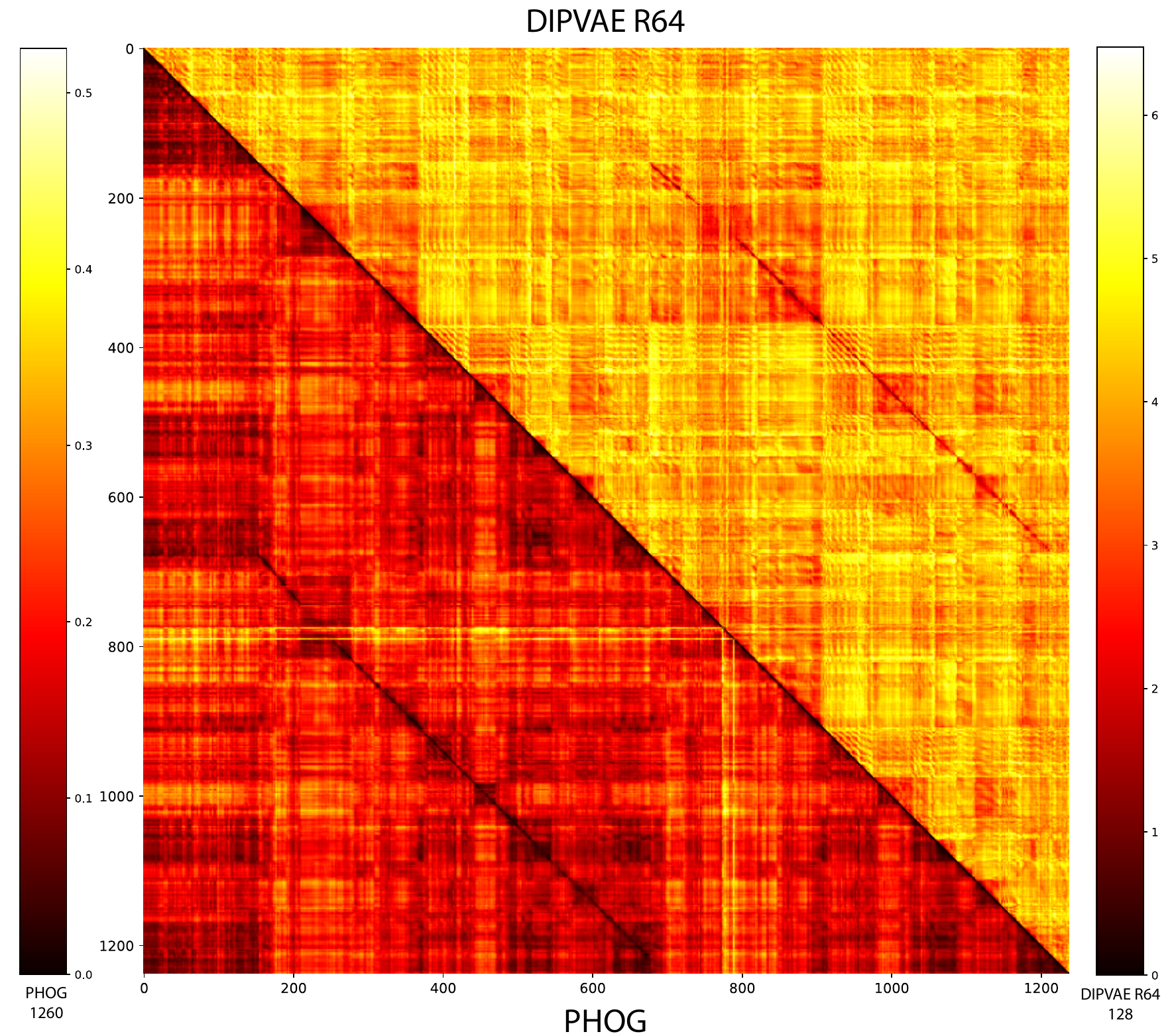}
            \caption{PHOG | DIPVAE R64}
        \end{subfigure}
        \begin{subfigure}[t]{0.46\columnwidth}
            \includegraphics[width=\columnwidth]{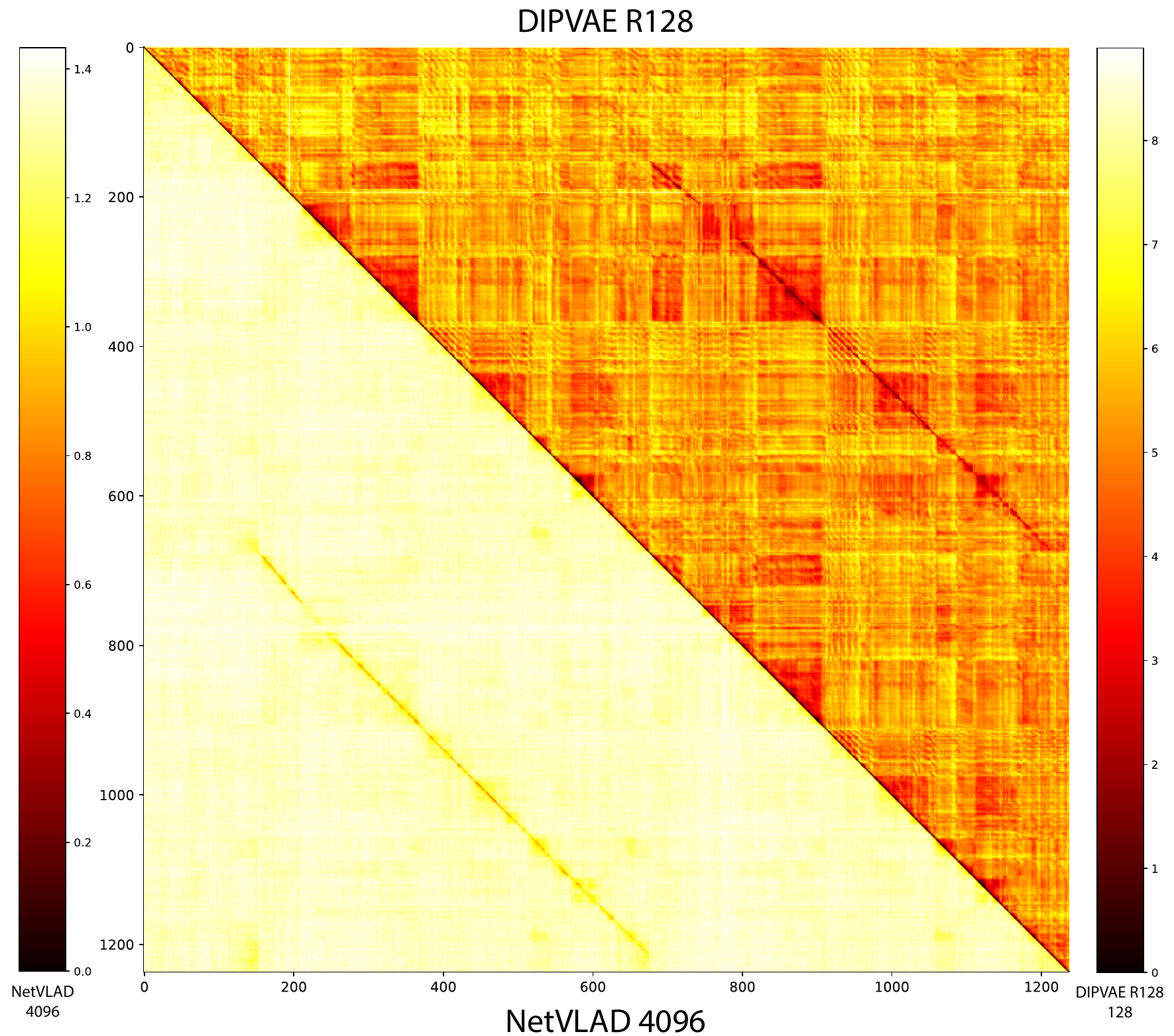}
            \caption{NetVLAD | DIPVAE R128}
        \end{subfigure}
    \end{subfigure}
    \vskip 8pt
    \begin{subfigure}[t]{\columnwidth}
        \centering
        \begin{subfigure}[t]{0.46\columnwidth}
            \includegraphics[width=\columnwidth]{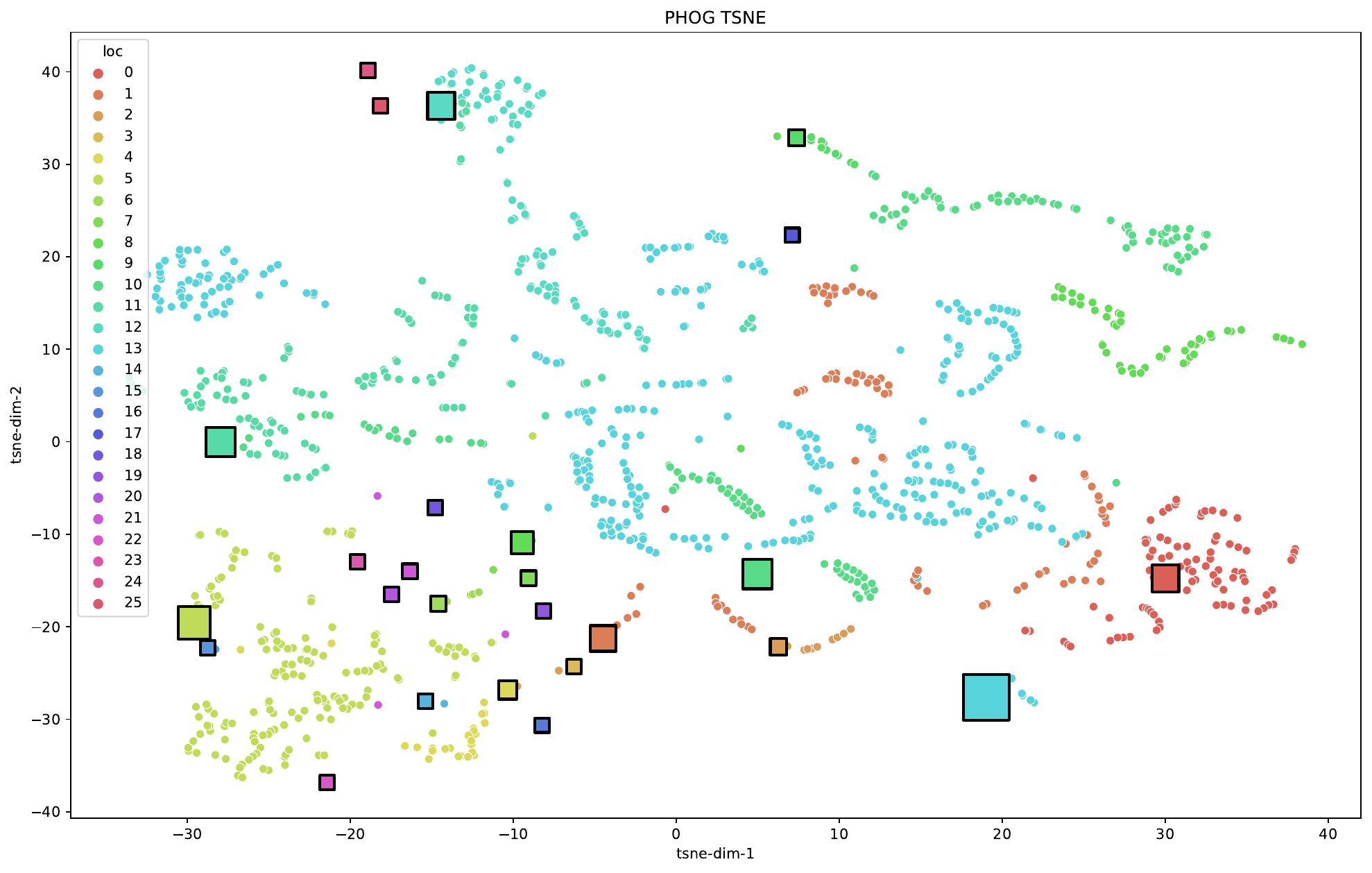}
            \caption{PHOG}
        \end{subfigure}
        \begin{subfigure}[t]{0.46\columnwidth}
            \includegraphics[width=\columnwidth]{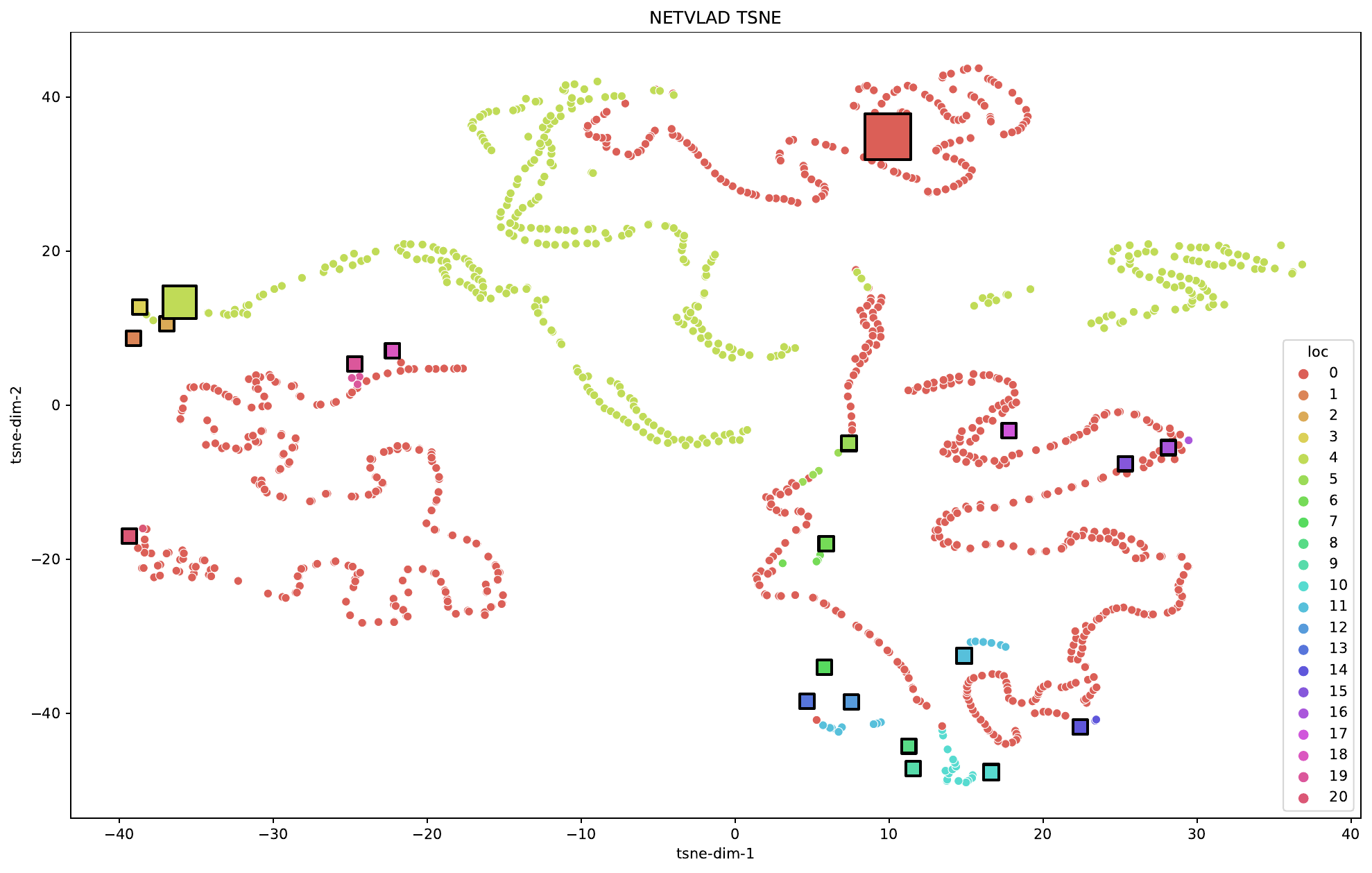}
            \caption{NetVLAD}
            \label{fig:tsne_plots_netvlad}
        \end{subfigure}
    \end{subfigure}
    \caption{Distance matrices (top row) and t-SNE plots (bottom row) of different global descriptors for City Centre dataset with 20-25 locations. To conserve space, we present the combined distance matrices (lower and upper triangular matrices) along with corresponding distance scales on the sides. In t-SNE plots, locations are shown in squares with its size linked to the number of images it has and its images are shown as dots of the same colour. Best viewed zoomed on a high resolution screen.}
    \label{fig:tsne_plots}
    \vskip -12pt
\end{figure}

To further substantiate this hypothesis, we conduct additional analysis utilizing distance matrices (inverse similarity matrices) and t-SNE (t-distributed Stochastic Neighbor Embedding) which are presented in \autoref{fig:tsne_plots}. An example of very smoothly changing but less distinctive nature exhibited by PHOG and an example of a less continuous but an overly distinctive nature exhibited by NetVLAD, are presented for City Centre dataset. Distance matrices of St Lucia dataset (21815 x 21815) are too large to interpret any meaningful information from, and hence, are not presented. PHOG exhibits an overabundance of similar descriptors and hence lacks distinctiveness, whereas NetVLAD has a scarcity of similar descriptors and hence lacks continuity. As such, the former leads to many false positive location matches, while the latter leads to poorly balanced locations (i.e. few locations having many too many images and/or many locations having a very few images), both resulting in decreased search efficiency, and thus, diminishing scalability. We also present the distance matrices of DIPVAE R64 and DIPVAE R128 for better comparison.

\begin{table}
\vskip 5pt
\centering
\begin{tabular}{lrr} \hline
\textbf{GDescriptor} & \textbf{n(Loc)} & \textbf{Runtime (s)} $\downarrow$ \\ \hline
PHOG            & 721                              & 27939.47                                           \\
LoST            & 654                              & 12041.02                                           \\
NetVLAD         & 624                              & 6753.36                                            \\
NetVLAD Cropped & 698                              & 8878.51                                            \\
DIPVAE R128     & 665                              & \textbf{2967.95}                  \\
DIPVAE R64      & 609                              & 3070.28                  \\ \hline
\end{tabular}
\caption{Runtime on St Lucia~\cite{stlucia_glover2010fab} dataset (17.6 km) for various descriptors with the number of locations produced corresponding to that run.}
\label{tab:st_lucia_runtime}
\vskip -12pt
\end{table}

The above results demonstrate the superiority of learned descriptors over handcraft descriptors, which we argue is due to their inherent continuity and distinctiveness. Overall, both NetVLAD and DIPVAE show flat runtimes and FPLC curves supporting their usage as global descriptors and demonstrating their scalability in systems that deal with longer trajectories. However, we note that in the longest track evaluated, St Lucia (17.6 km, 21.8K images), DIPVAE (both R128 and R64 variants) shows significantly lower runtimes as shown in \autoref{tab:st_lucia_runtime}, while being very efficient with a compact embedding length of 128 and substantially faster compute time as noted in \Cref{tab:eval_candidates}. We note that the LoST descriptor that encodes semantic information resulted in a longer runtime with more FPLC compared to its learned descriptor counterparts. We hypothesize that this is a result of the weaker discriminability arising due to ambiguous semantic information present across multiple regions. 

\section{Conclusion}
\label{Conclusion}
We have extended HTMap~\cite{htmap_Garcia-Fidalgo2017} making improvements and incorporating learned global descriptors into the framework. We perform a comprehensive evaluation of various global descriptors on hierarchical topological mapping and present results of recall at 100\% precision, total runtimes and false positive loop closure location candidates (FPLC). All of the global descriptors we compared yield similar overall recall, however, show crucial differences in runtimes. Based on our empirical analysis of multiple runs, we have identified that continuity and distinctiveness are crucial characteristics for an optimal global descriptor that enable efficient and scalable hierarchical mapping. Additionally, we have presented a methodology for quantifying and contrasting these characteristics. Our study demonstrates that the global descriptor based on unsupervised learned Variational Autoencoder (VAE) excels in these characteristics and achieves significantly lower runtime. Consequently, on the longest track, St Lucia (17.6 km), we observed that DIPVAE outperforms other global descriptors significantly, running upto 2.3x faster than the next best global descriptor, NetVLAD, while upto 9.5x faster than PHOG, while also being compact with an embedding length of 128.
In the future we intend to extend our analysis to more challenging datasets including more extreme variation in weather, seasons, daylight, etc. We expect the performance of the learned descriptors to be more pronounced in these settings.

\section*{Acknowledgements}
This research was supported 
with the financial support of the
the \href{https://www.sfi.ie}{Science Foundation Ireland} (SFI) grants 13/RC/2094 and 16/RI/3399.
For the purpose of Open Access, the author has applied a CC BY public copyright licence to any Author Accepted Manuscript version arising from this submission. 
The opinions, findings and conclusions or recommendations expressed in this material are those of the author(s) and do not necessarily reflect the views of SFI and/or Lero.

\bibliographystyle{ieeetr}
\bibliography{root}

\end{document}